\documentclass{article}

\usepackage[final]{nips_2017}

\usepackage[utf8]{inputenc} 
\usepackage[T1]{fontenc}    
\usepackage[colorlinks=true,linkcolor=black,anchorcolor=black,citecolor=black,filecolor=black,menucolor=black,runcolor=black,urlcolor=black]{hyperref}       
\usepackage{url}            
\usepackage{booktabs}       
\usepackage{amsfonts}       
\usepackage{nicefrac}       
\usepackage{microtype}      

\usepackage{amsmath}
\usepackage[pdftex]{graphicx}
\usepackage{subcaption} 

\title{Unsupervised patient representations from clinical notes with interpretable classification decisions}

\author{
Madhumita Sushil\textsuperscript{1,2,3}, Simon \v{S}uster\textsuperscript{2,4}, Kim Luyckx\textsuperscript{1,5}, Walter Daelemans\textsuperscript{2,4} \\
\textsuperscript{1}Antwerp University Hospital, Belgium \\
\textsuperscript{2}CLiPS Research Center, University of Antwerp, Belgium \\
\textsuperscript{3}\texttt{firstname.lastname@outlook.com} \\
\textsuperscript{4}\texttt{firstname.lastname@uantwerpen.be} \\ 
\textsuperscript{5}\texttt{firstname.lastname@uza.be}
  }

\begin{document}

\maketitle

\begin{abstract}

We have two main contributions in this work:
1. We explore the usage of a stacked denoising autoencoder, and a paragraph vector model to learn task-independent dense patient representations directly from clinical notes. We evaluate these representations by using them as features in multiple supervised setups, and compare their performance with those of sparse representations. 2. To understand and interpret the representations, we explore the best encoded features within the patient representations obtained from the autoencoder model. Further, we calculate the significance of the input features of the trained classifiers when we use these pretrained representations as input.
 
\end{abstract}

\section{Introduction}
\label{sec:intro}

Representation learning techniques have been used extensively within and outside the clinical domain to learn the semantics of words, phrases, and documents~\citep{DBLP:conf/acl/BaroniDK14,DBLP:journals/corr/LiuCJY16}. We apply such representation learning techniques to create a patient semantic space by learning vector representations at the patient level. In a patient semantic space, ``similar" patients should have similar vectors. Patient similarity metrics are widely used in several applications to assist clinical staff. Some examples are finding similar patients for rare diseases~\citep{garcelon2017finding}, identification of patient cohorts for disease subgroups~\citep{li2015identification}, providing personalized treatments~\citep{zhang2014towards,wang2015electronic}, and predictive modeling tasks such as patient prognosis~\citep{gottlieb2013method,wang2012medical} and risk factor identification~\citep{ng2015personalized}. When patient similarity is calculated as an ontology-guided distance between specific structured properties of patients such as diseases and treatments, it represents patient relationships corresponding to those properties. However, when the similarity measure is fuzzy, the different properties that influence the similarity value are unknown. We aim to capture patient similarity on multiple dimensions, such as complaints, diagnoses, procedures performed, etc., which would encapsulate a holistic view of the patients. 

In this work, we create dense patient representations that are transferable across tasks from clinical notes in the freely available MIMIC-III database~\citep{johnson2016mimic}. We focus on different techniques for creating patient representations using only textual data. We explore the usage of two neural representation learning architectures---a stacked denoising autoencoder~\citep{vincent2010stacked}, and a paragraph vector architecture ~\citep{le2014distributed}---for unsupervised learning. We evaluate the quality of the learned representations through multiple supervised tasks.

Dense representations can capture semantics, but at a loss of interpretability. We take a step towards bridging this gap by proposing different techniques to interpret the information encoded in the patient vectors obtained during the unsupervised learning phase, and to extract the features that most influence the classification output when they are used as input.

\section{Learning Patient Representations}
\label{sec:methods}

\paragraph{Stacked denoising autoencoder:}

\citet{miotto2016deep} used a stacked denoising autoencoder (SDAE)~\citep{vincent2010stacked} to learn patient representations for disease prognosis using structured patient data combined with probabilistic topic models obtained from unstructured data. \citet{suresh2016use} used a sequence-to-sequence autoencoder to generate patient phenotypes using structured data. Given the success of these models, we explore the use of an SDAE for task-independent patient representation from unstructured data forgoing the use of intermediate techniques like topic modeling. 
%
%

We sequentially train every layer of an SDAE as an independent denoising autoencoder to reconstruct the hidden layer output of the previous autoencoder from a corresponding corrupted version. We use the hidden layer value of the final autoencoder as the dense patient representations $R$. We use the sigmoid activation function for the encoders, and the linear activation function for the decoders. We train the network to minimize the mean squared reconstruction error using the RMSProp optimizer~\citep{tieleman2012lecture}. After a randomized hyperparameter search~\citep{bergstra2012random}, we obtain a 1-layer SDAE with 800 hidden units and 5\% dropout noise.

\paragraph{Paragraph vector:}
Doc2vec, or `Paragraph Vector'~\citep{le2014distributed}, learns dense fixed-length representations of variable length texts such as paragraphs and documents. It supports two algorithms---a distributed bag-of-words (DBOW) algorithm, and a distributed memory (DM) algorithm. 
We use the DBOW algorithm for 5 iterations, with a window size of 3, a minimum frequency threshold of 10, and 5 negative samples per positive sample to train 300 dimensional patient vectors. We determined these settings also using the randomized hyperparameter search.

\section{Feature extraction}
\label{sec:feat_ext}

When statistical models are deployed for clinical decision support, it is crucial to understand the features that influence the model output. A ranked list of the most influential features can assist such understanding, 
%
%
while facilitating error analysis, and exploratory analysis when unexpected features are ranked high. We propose two techniques to achieve model interpretability. First, to estimate how well the individual features are encoded in the patient vectors learned through the SDAE, we calculate the \textbf{squared reconstruction error} of the input features in the first layer of the pretrained autoencoder, averaged across all the training instances. Next, we extend the work by~\citet{engelbrecht1998feature} and use \textbf{sensitivity analysis} to calculate the significance of the original input words for different classification tasks for a selected set of instances, when the task-independent dense patient representations $R$ are first generated using the SDAE, and $R$ is then used as the input to the classifiers. This technique is transferable to the doc2vec representations and we plan to extend it in future. Given an input $R$ to the classifier corresponding to the original inputs $z$ to the SDAE model, the significance of the $i$th input feature, $\phi_{z_i}$ is defined as the maximum significance of the input feature $i$ across all the $K$ output units ($o$) of the classifier with respect to the $N$ instances:

\begin{align*}
\phi_{z_i} = \max_{k = 1...K}\{S_{oz,ki}\} 
&\text{,\;where\;}
S_{oz,ki} = \sqrt{\sum\limits_{j=1}^N [S_{oz,ki}^{(j)}]^2 * N^{-1} } .
\end{align*}


$S_{oz,ki}^{(j)}$ is the sensitivity of the $k$th output unit of the classifier w.r.t the $i$th input feature of the SDAE for an instance $j$: 
$$ S_{oz,ki}^{(j)} = \frac{\partial o_k^{(j)}}{\partial z_i^{(j)}} =   \frac{\partial o_k^{(j)}}{\partial R_i^{(j)}} *  \frac{\partial R_i^{(j)}}{\partial z_i^{(j)}} . $$


\section{Dataset construction and preprocessing}
\label{sec:dataset}
We retrieve a set of adult patients ($\geq$18 years) with one hospital admission and at least one associated note (excluding discharge reports) from the MIMIC-III database~\citep{johnson2016mimic}. We split it into a set of 24,650 patients for training, and 3,081 patients each for validation and testing. 
We represent patients with a concatenation of all their non-discharge notes. We tokenize the data using the Ucto tokenizer~\citep{van2012ucto} and lowercase it. To obtain patient representations using the SDAE, we replace the numbers, and certain time and measurement mentions with special tokens. We remove the punctuations, and the terms with frequency $<5$. We use a bag-of-words (BoW) with their TF-IDF scores as features,
to obtain a vocabulary size of 71,001. We also conducted the experiments with a bag-of-medical-concepts feature set, but they performed consistently worse. To train the doc2vec models, we remove the numbers, and the tokens matching time and measurement patterns (determined from the initial validation set results), and get a vocabulary size of 48,950.\looseness=-1 
%
%

\section{Evaluation}

\subsection{Task description}
We use the dense patient representations $R$ as the input features to train feedforward neural network classifiers for the following independent tasks: 
binary prediction of patient mortality during the hospital stay (13.14\%), within 30 days of discharge (3.85\%), or within 1 year of discharge (12.19\%); prediction of the 20 generic diagnostic categories, and the 18 generic procedural categories as encoded in the ICD-9-CM database~\citep{world2004international}, corresponding to the most relevant diagnostic and procedural codes for a patient (the majority classes are 40.2\% and 38.9\% respectively); and gender prediction---male (56.87\%) or female (43.13\%). We evaluate the models using the area under the ROC curve for patient death for the mortality tasks, and the weighted F-score for the others, to correct for class imbalance. We minimize the categorical cross-entropy error using the RMSProp optimizer, and determine the hyperparameters using randomized search.

\subsection{Results and Discussion}
\label{sec:results}

\begin{table}[t]
  \caption{Classification results on different tasks using the BoW features, the SDAE and the doc2vec patient representations, and on concatenating the two dense representations (with $\kappa$ score).}
  \label{results_bow}
  \centering
  \resizebox{\textwidth}{!}{
  \begin{tabular}{lrrr|rrr}
    \toprule
     Approach & In\_hosp  & 30\_days & 1\_year &  Pri\_diag\_cat & Pri\_proc\_cat & Gender \\
    \midrule
   BoW & 94.57 & 59.49 & 79.42 & 70.16 & 73.66 & 98.47\\
   SDAE & 91.94 & 79.65 & 79.80 & 65.00 & 67.46 & 87.75 \\
  doc2vec & 91.95 & 76.80 & 81.34 & 68.07 & 65.83 & 97.70 \\
 ($\kappa$) SDAE + doc2vec & (58.65) 93.83 & (00.00) 81.13 & (15.81) 83.02 & (64.38) 67.88 & (58.91) 70.30 & (72.00) 97.47 \\
    \bottomrule
  \end{tabular}}
\end{table}

In Table~\ref{results_bow}, we compare the classification performance on using the dense patient representations obtained from the SDAE and the doc2vec models as the input features for all the tasks, compared to the BoW sparse features. We analyze the agreement between the SDAE and the doc2vec model outputs by calculating Cohen's $\kappa$ score~\citep{cohen1960coefficient} between them on the validation set. We find that the agreement scores are not high, which may indicate that the models learn complimentary information. We then concatenate the two dense representations to analyze model complementarity.

Our main finding is that all the dense representation techniques significantly outperform\footnote{All the statistical significance scores were calculated using the two-tailed pairwise approximate randomization test~\citep{noreen1989computer} with a significance level of 0.05 before the Bonferroni correction for 36 hypotheses.}
the baseline for 30 days mortality prediction. However, although we see a large numerical improvement over the baseline on using the dense representations for 1 year mortality prediction, the differences are not statistically significant. We believe that the poor performance of the BoW model for 30 days mortality prediction may be due to the low number of positive instances, and that generalization assists feature identification in such cases. \citet{grnarova2016neural} have previously shown significant improvements for these tasks on using a 2-level convolutional neural network as compared to the doc2vec vectors used in linear support vector machines. However, our results are not directly comparable because we use different data subsets, and non-linear neural classifiers with the doc2vec representations. The sparse inputs perform better than the SDAE representations for all the other tasks, and better than the doc2vec representations for in-hospital mortality and primary procedural category prediction. One probable reason is that the best predictors for the other tasks are the direct lexical mentions in the notes, which makes the BoW model a very strong baseline. Examples of such features obtained using the $\chi^2$ feature analysis are `autopsy', `expired', `funeral', and `unresponsive' for in-hospital mortality prediction, and `himself', `herself', `ovarian', and `testicular' for gender prediction.

The concatenation of the vectors learned by both models is not statistically different from the sparse representations under the given significance level for any task except 30 days mortality prediction, where the concatenation is better. This ensemble model significantly outperforms both individual models for primary procedural category prediction. For primary diagnostic category and gender prediction, the ensemble model is significantly better than the SDAE model, but not the doc2vec model. In these cases, there is no significant difference between the doc2vec and the BoW models. Hence, we observe that the concatenation helps in some cases and we recommend combining the two dense representations for unknown tasks. The doc2vec model uses a local context window in a log-linear classifier, whereas the SDAE model uses only the global context information and non-linear encoding layers. This may be one of the factors governing the differences between the two techniques. \looseness=-1

Furthermore, we rank the features according to their mean squared reconstruction error when we pretrain the patient representations using the SDAE. We observe that infrequent terms such as spelling errors are reconstructed very well, as opposed to the frequent features in the dataset. To check for a correlation between this error and the feature frequency, we calculate the Spearman's rank correlation coefficient~\citep{kokoska2000crc} between the two parameters, and obtain a value of 0.8738. We believe that this behavior may be due to the high entropy of the frequent terms. \citet{jocombining} also obtain misspellings and rare words as the top features when they use recurrent neural networks for patient mortality prediction in the MIMIC-III dataset.

\begin{table}[t]
  \caption{The most significant features for the classifiers for one test instance each when the SDAE representations are used as the input. The true classes are `patient death' for the mortality tasks (a common instance for 30 days and 1 year mortality prediction), and `diseases of the digestive system', `operations on the digestive system', and `male' respectively for a common patient for the other tasks. }
  \label{feat_imp}
  \centering
  \resizebox{\textwidth}{!}{
  \begin{tabular}{llllll}
    \toprule
     In\_hosp & 30\_days & 1\_year & Pri\_diag\_cat & Pri\_proc\_cat & Gender \\
    \midrule
     vasopressin & leaflet & magnevist & numeric\_val & numeric\_val & woman \\
     pressors & structurally & signal & previous & no & female \\
     focused & pacemaker & decisions & rhythm & of & she\\
     dnr & sda & periventricular & no & enzymes & man \\
     dopamine & periventricular & embolus & flexure & extubated & he \\
     acidosis & excursion & underestimated & dementia & rhythm & male\\
     levophed & non-coronary & calcified & brbpr  & and & her \\
     pressor & dosages & screws & of  & the & his \\
     cvvhd & microvascular & rib & sinus & vent & wife \\
     cvvh & left-sided & shadowing & for & uncal & uterus \\
     emergency & chronic & gadolinium & to & mso & him \\
     pneumatosis & extubation & mri & tracing & to & urinal \\
    \bottomrule
  \end{tabular}}
\end{table}

In Table~\ref{feat_imp}, we list the most significant features for the model outputs for one instance each in the test set. We find that the classifiers give high importance to sensible frequent features for most of the tasks, although the SDAE reconstructs the low frequent terms better during the pretraining phase. There is a minimal overlap between the sets of important features for different tasks. This shows that $R$ is task-independent, and that the classifiers can identify task-specific important information when they are trained for a specific task. 

\section{Conclusions}
We have shown that the dense patient representations significantly improve the classification performance for 30 days mortality prediction, a task where we are confronted with a very low proportion of positive instances. For the other tasks, this advantage is not visible. Moreover, we have shown that a combination of the stacked denoising autoencoder and the doc2vec representations improves over the individual models for some tasks, without any harm for the others tasks. Furthermore, during feature analysis, we have found that the frequent terms are not encoded well during the pretraining phase of the stacked denoising autoencoder. However, when we use these pretrained vectors as the input, sensible frequent features are selected as the most significant features for the classification tasks.

\subsubsection*{Acknowledgments}
This research was carried out within the Accumulate SBO project (\url{www.accumulate.be}), funded by the government agency Flanders Innovation \& Entrepreneurship (VLAIO), Belgium [grant number 150056].

\bibliographystyle{apalike}
\bibliography{patient_rep}

\end{document}